\documentclass[letterpaper]{article} %
\usepackage{aaai25}
\usepackage{times}  %
\usepackage{helvet}  %
\usepackage{courier}  %
\usepackage[hyphens]{url}  %
\usepackage{graphicx} %
\usepackage{natbib}  %
\usepackage{caption} %
\usepackage{amsfonts, dsfont}
\usepackage{multirow}
\usepackage{booktabs}
\usepackage[table]{xcolor}
\usepackage{subcaption}
\usepackage{algorithm}
\usepackage{algorithmic}

\usepackage{newfloat}
\usepackage{listings}
\DeclareCaptionStyle{ruled}{labelfont=normalfont,labelsep=colon,strut=off} %
\floatstyle{ruled}
\newfloat{listing}{tb}{lst}{}
\floatname{listing}{Listing}
\title{Harnessing Large Language and Vision-Language Models for Robust Out-of-Distribution Detection}
\author{Pei-Kang Lee$^1$, Jun-Cheng Chen$^2$, Ja-Ling Wu$^1$}
\affiliations{1. National Taiwan University\\
2. Academia Sinica}

\usepackage{bibentry}

\begin{document}

\maketitle

\begin{abstract}

Out-of-distribution (OOD) detection has seen significant advancements with zero-shot approaches by leveraging the powerful Vision-Language Models (VLMs) such as CLIP. However, prior research works have predominantly focused on enhancing Far-OOD performance, while potentially compromising Near-OOD efficacy, as observed from our pilot study. To address this issue, we propose a novel strategy to enhance zero-shot OOD detection performances for both Far-OOD and Near-OOD scenarios by innovatively harnessing Large Language Models (LLMs) and VLMs. Our approach first exploit an LLM to generate superclasses of the ID labels and their corresponding background descriptions followed by feature extraction using CLIP. We then isolate the core semantic features for ID data by subtracting background features from the superclass features. The refined representation facilitates the selection of more appropriate negative labels for OOD data from a comprehensive candidate label set of WordNet, thereby enhancing the performance of zero-shot OOD detection in both scenarios. Furthermore, we introduce novel few-shot prompt tuning and visual prompt tuning to adapt the proposed framework to better align with the target distribution. Experimental results demonstrate that the proposed approach consistently outperforms current state-of-the-art methods across multiple benchmarks, with an improvement of up to 2.9$\%$ in AUROC and a reduction of up to 12.6$\%$ in FPR95. Additionally, our method exhibits superior robustness against covariate shift across different domains, further highlighting its effectiveness in real-world scenarios.
\end{abstract}

\section{Introduction}
Out-of-distribution (OOD) detection is a critical and active research area that aims to identify whether input samples belong to the distribution of the training data used in machine learning (ML) models. This capability is crucial for preventing unpredictable outputs from ML models when faced with unseen inputs in real-world applications~\cite{ulmer2020trust}.

Traditionally, OOD detection approaches have primarily relied on Convolutional Neural Networks (CNN), focusing on analyzing visual features \cite{lee2018simple, sastry2020detecting}, or logit spaces \cite{liu2020energy}. However, these methods are usually limited by either their limited model capacities or their reliance on visual information alone. The emergence of powerful Vision-Language Models (VLMs) like Contrastive Language-Image Pretraining (CLIP) \cite{radford2021learning}, which is trained with 400 million image-text pairs, has opened new avenues for tackling OOD detection by utilizing their rich semantic information in both visual and label spaces. Recent studies \cite{esmaeilpour2022zero, ming2022delving, jiang2024neglabel} have explored using CLIP for zero-shot OOD detection, demonstrating promising results.
For instance, one of the representative CLIP-based zero-shot OOD detectors, NegLabel~\cite{jiang2024neglabel}, augments the class labels for ID data with the selected negative labels from the noun and adjective lexname categories of WordNet~\cite{fellbaum1998wordnet} as the proxies for OOD classes to enhance the performance.  
However, these works mainly focus on studying Far-OOD scenarios where the distributions of in-distribution (ID) and OOD data are distant. Additionally, while zero-shot OOD detection methods using CLIP have shown great potential, they may struggle to capture the nuances and specific characteristics of downstream tasks \cite{wei2023improving}.
Therefore, when the distributions of ID and OOD data are similar for the Near-OOD scenarios, it is still a challenging problem requiring more discriminative and detailed information.\\
\indent To tackle this challenge, we propose a novel approach to further enhance the power of CLIP for zero-shot OOD detection in both Near-OOD and Far-OOD scenarios by expanding the selection space of OOD candidate labels and refining the semantic space of ID labels. To achieve the expansion, we employ a comprehensive set of WordNet categories for candidate OOD label selection by performing a similar negative-mining algorithm as NegLabel without any candidate label filtering. As we find from our experimental results (c.f. Table~\ref{tab:ablation}) when NegLabel performs negative-mining for appropriate negative labels, it filters out specific candidate labels and leads to suboptimal performance.
Furthermore, to refine the semantic space of ID labels, we leverage Large Language Models (LLMs) to generate superclass labels and background descriptions to create a proxy for selecting negative labels. This process involves initially broadening the ID semantic space through superclass identification with LLM followed by a subtraction step of background information to refine the space. The refined ID space enables negative-mining to select more representative negative labels from the expanded selection space of OOD candidate labels. With our proposed hierarchical strategy, our zero-shot approach outperforms not only current state-of-the-art zero-shot but also CLIP-based training methods, demonstrating the effectiveness of combining VLMs with LLMs for OOD detection.

To further enhance the performance of our zero-shot OOD method, we adapt CLIP to the target distributions by our two-phase training method with few-shot prompt tuning (PT) and visual prompt tuning (VPT). Moreover, to address the practical challenge of collecting unlabeled OOD samples for training, we adopt the same approach as ID-like \cite{bai2024idlike} by leveraging less label-relevant portions of ID samples as OOD data to create a robust OOD proxy dataset while maintaining the inherent characteristics of the ID domain. Together with this method, it allows us to benefit from outlier exposure during few-shot training without relying on external OOD data. Our proposed few-shot method has been extensively evaluated on multiple benchmarks with state-of-the-art performances, including the ImageNet-1K OOD benchmark \cite{huang2021importance} and the challenging OpenOOD V1.5 full-spectrum benchmark \cite{zhang2023openood, yang2022fsood}. Our method significantly outperforms existing approaches, achieving the average AUROC of 97.67\% on the ImageNet-1K OOD benchmark and the average AUROC of 83.04\% on the challenging OpenOOD V1.5 Near-OOD full-spectrum benchmark.
In summary, our key contributions are as follows:
\begin{itemize}
\item Improved zero-shot OOD detection: We introduce a hierarchical strategy using LLMs to generate superclass labels and background descriptions, creating more representative negative labels to enhance the zero-shot OOD detection performance.
\item Novel few-shot OOD detection method: We propose a new two-phase few-shot learning framework that exploits our proposed PT and VPT to adapt CLIP to the target distributions. Together with the ID-like auxiliary OOD data generation, the proposed approach achieves state-of-the-art OOD detection performance.
\end{itemize}
\begin{figure*}
\centering
\includegraphics[width=\linewidth]{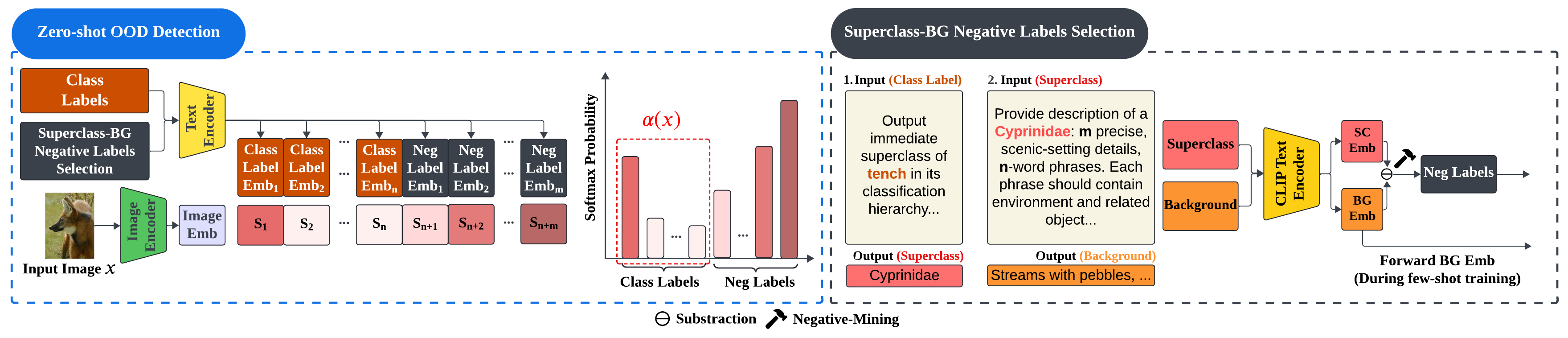}
\caption{Illustration of the proposed zero-shot OOD detection and Superclass-BG negative label selection. We harness the capabilities of LLMs to select more representative negative labels. 
}\label{fig:zero_shot_overview}
\end{figure*}
\begin{figure*}
\centering
\includegraphics[width=\linewidth]{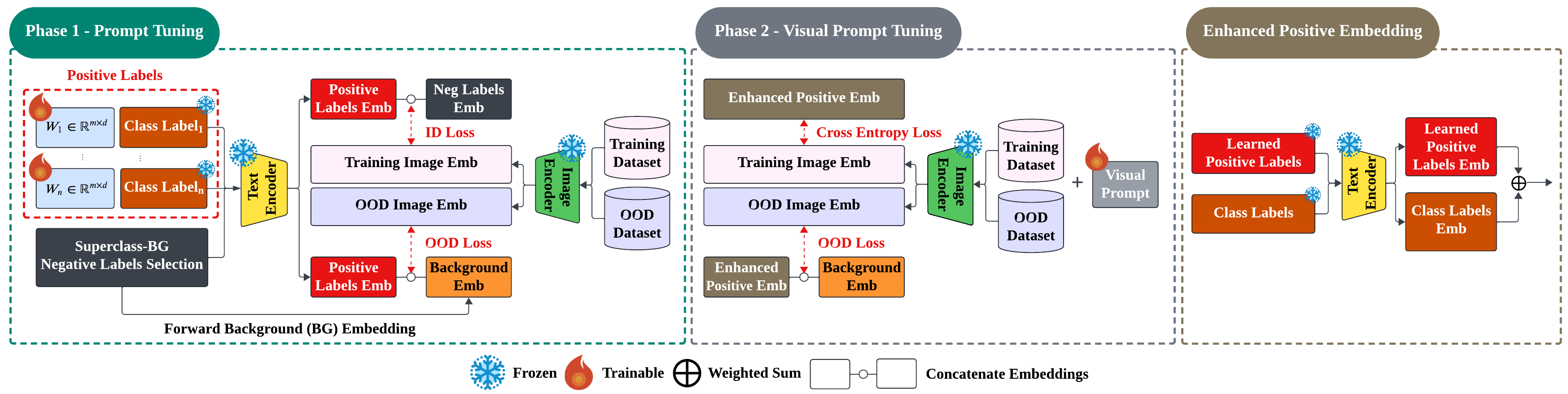}
\caption{Illustration of the proposed few-shot learning framework. The training dataset consists of a few-shot sample from the ImageNet-1K training set, while the OOD dataset is generated using an ID-like~\cite{bai2024idlike} approach. Here, the term \textit{positive labels} refer to the class labels with learnable prompt. }\label{fig:few_shot_overview}
\end{figure*}
\section{Related Work}
While OOD detection has a rich history in computer vision, our work focuses on leveraging VLMs for OOD detection. We briefly review the relevant works as follows. 

\subsubsection{CLIP-based OOD Detection.}
VLMs, such as CLIP, possess formidable zero-shot capabilities and robustness against distribution shifts. Recognizing these strengths, an increasing number of researchers are focusing on harnessing the power of VLMs to enhance OOD detection~\cite{Miyai2024GeneralizedOOD}. ZOC~\cite{esmaeilpour2022zero} trains an image descriptor to obtain descriptions of images, using these additional descriptions as supplementary OOD labels. MCM \cite{ming2022delving} uses class labels as prototypes and exploits the uniformity property of contrastive learning to enhance the discrepancy between ID and OOD samples through softmax rescaling. Another research direction aims to teach CLIP the logic of contrary semantics, namely the concept of \textit{no}. CLIPN \cite{wang2023clipn} fine-tunes using the CC12M \cite{changpinyo2021cc12m} dataset, a \textit{no} text encoder, and \textit{no} prompts to teach CLIP the logic of \textit{no}. Meanwhile, LSN \cite{nie2024outofdistribution} leverages ID data to learn this \textit{no} logic. NegPrompt \cite{li2024learning} learns a set of negative prompts that contain semantics opposite but related to the ID labels. Furthermore, employing additional OOD labels to make ID images gravitate towards ID labels and away from selected OOD labels is also a popular research direction. TOE \cite{park2023powerfulness} demonstrates the utility of textual outliers, achieving promising results through training methods similar to image outlier exposure. ID-like \cite{bai2024idlike} uses random cropping to identify images most closely aligned and unrelated to the corresponding label, aiming to align textual prompts and learnable OOD labels with these images. NegLabel \cite{jiang2024neglabel}, on the other hand, employs negative-mining from a large word corpus to identify labels semantically distant from ID semantic space.

\section{Preliminaries}

\subsubsection{Zero-shot Classification with CLIP.}
CLIP consists of a text encoder $\mathcal{T}: t \rightarrow \mathbb{R}^d$ and an image encoder $\mathcal{I}: x \rightarrow \mathbb{R}^d$, mapping text $t$ and image $x$ to $d$-dimensional feature vectors. For zero-shot classification with label set $C = \{c_1, ..., c_m\}$, the vectors $t_j$ are constructed using prompts ``a photo of a $\langle c_j \rangle$''. The similarity score between image features $\mathcal{I}(x)$ and text features $\mathcal{T}(t_j)$ is computed as the cosine similarity of $\mathcal{T}(t_j)$ and $\mathcal{I}(x)$. The predicted probability for class $c_j$ is computed using the softmax function over the similarity scores, with a temperature parameter $\tau$.

\subsubsection{Prompt Tuning and Visual Prompt Tuning}
Prompt tuning (PT) methods like CoOp \cite{zhou2022learning} initialize learnable prompt tensors $t = [V]_1...[V]_L[\mathrm{CLASS}]$, where $[V]_l$ are learnable vectors. These prompts are optimized using few-shot examples to improve the performance on target distributions. The concept of the visual prompt tuning (VPT) \cite{jia2022vpt}, akin to that used in CoOp, involves the insertion of fixed-length learnable vectors between the image patch embeddings and the class token.

\subsubsection{Out-of-Distribution Detection.}
OOD detection aims to construct a binary classifier $G(x)$:
\footnotesize
\begin{equation}
G(x) = \Bigl\{
\begin{array}{ll}
\mathrm{ID},  & \mathrm{if } \; \alpha(x) \geq \gamma \\
\mathrm{OOD}, & \mathrm{if } \; \alpha(x) < \gamma
\end{array}
\end{equation}
\normalsize
where $\alpha(x)$ is a score function and $\gamma$ is a threshold.

\section{Methodology}
In this section, we describe the detailed of the proposed zero-shot and few-shot training OOD detection as follows. 

\subsection{Zero-Shot OOD Detection}
As shown in Figure~\ref{fig:zero_shot_overview}, we perform similar CLIP-based zero-shot classification for zero-shot OOD detection. Moreover, we augment the class labels for ID data with negative labels for OOD classes from a comprehensive set of WordNet categories using our proposed LLM-based Superclass-\textbf{B}ack\textbf{G}round (Superclass-BG) negative label selection algorithm, as detailed below (i.e., we do not apply any candidate label filtering during the process). Additionally, the selected negative labels are partitioned into $K$ distinct groups following the grouping strategy in NegLabel \cite{jiang2024neglabel}.

\subsubsection{LLM-based Superclass-BG Negative Label Selection.}
Let $C = \{c_1, c_2, ..., c_m\}$ denote the set of class labels, $SC = \{sc_1, sc_2, ..., sc_n\}$ the set of superclass labels where $SC$ is generated from $C$ through LLMs, $n \leq m$, and $N = \{n_1, n_2, ..., n_p\}$ the set of candidate labels. Define \(S(u, v)\) as the measure of cosine similarity between the embedding \(u\) and \(v\). The negative-mining algorithm introduced in NegLabel calculates $S(\mathcal{T}(n_i), \mathcal{T}(c_j))$ for all $n_i \in N$ and $c_j \in C$, where $\mathcal{T}$ denotes the CLIP text encoder. Defining $\mathrm{Score}(n_i)$ as the $q$-quantile of $\{S(\mathcal{T}(n_i), \mathcal{T}(c_j)) | c_j \in C\}$. This approach of using a quantile, rather than a minimum distance or average similarity, provides robustness against outlier ID labels by considering the overall distribution of similarities. The negative labels are determined by selecting the labels corresponding to the lowest $p$ percentage of $\mathrm{Score}(n_i)$ values, where $p$ is a predefined threshold. Our proposed method extends this approach by leveraging superclass labels generated by LLMs. We calculate $\mathrm{Score}'(n_i)$ as the $q$-quantile of $\{S(\mathcal{T}(n_i), \mathcal{T}(sc_j)) | sc_j \in SC\}$. This superclass-based approach offers an improved representation of high similarity across a broader range of concepts, as each $sc_j$ encapsulates multiple related classes, mitigating the influence of semantically similar class clusters on the quantile calculation.
We leverage the semantic comprehension capability of CLIP \cite{Goh2021MultimodalNeurons} to further refine the semantic representation. For each superclass label $sc_j$, we also generate a background description $BG(sc_j)$ using LLMs. We then compute adjusted superclass label embeddings as $\mathcal{T}'(sc_j) = \mathcal{T}(sc_j) - \mathcal{T}(BG(sc_j))$. This subtraction aims to isolate the core semantic content of each superclass label by removing general background features. The rationale behind this additional step is twofold. First, it addresses potential inaccuracies in LLM-generated superclass labels, which may not perfectly align with optimal taxonomies. Second, it focuses the embeddings on distinguishing features of the superclass labels, potentially reducing semantic overlap.

\subsection{Two-Phase Training for Few-shot OOD Detection}
Our proposed method is structured into two distinct phases. In the initial phase, we focus exclusively on prompt tuning for class labels while maintaining all other model components fixed. The second phase builds upon the results of the first. We fix the tuned prompts obtained from the previous stage and proceed with VPT. The overview is illustrated in Figure~\ref{fig:few_shot_overview}. To derive the OOD datasets from few-shot training samples, we employ an ID-like~\cite{bai2024idlike} approach. Multiple cropped images are initially extracted from each training sample via random cropping. Subsequently, these cropped images are bifurcated based on their cosine similarity to the corresponding class label. Images exhibiting lower similarity are allocated to the OOD dataset. This approach enables the creation of a robust OOD proxy dataset while maintaining the inherent characteristics of the ID domain.
\subsubsection{Phase-1 Prompt Tuning.}
To facilitate subsequent discussions, we introduce the term \textit{positive labels}, denoted as $P$, to refer to the class labels with learnable prompts while all other components remain fixed. Let $\mathcal{D}_{\mathrm{train}}$ be the few-shot training dataset, $\mathcal{D}_{\mathrm{ID}}$ be the ID dataset and $\mathcal{D}_{\mathrm{OOD}}$ be the OOD dataset. The dissimilarity measure between embedding $u$ and $v$ is given by \(D(u, v) = 1-S(u, v)\) . Let \(C\) be the set of class labels. Let \(x\) be the input image from the few-shot training dataset, \(y\) be the corresponding ground-truth label and $\tau$ be the temperature. During the prompt tuning phase, the optimization process incorporates two distinct loss functions: the ID loss and the OOD loss. The ID loss is formulated in Equation~\ref{eq:PT_ID1} and~\ref{eq:PT_ID2}.
\footnotesize
\begin{equation}
\mathcal{L}_{\mathrm{ID}} = \max_k \; \mathbb{E}_{(x,y)\sim \mathcal{D}_{\mathrm{train}}}
\big[\mathcal{F}_k(x, y)\big],\label{eq:PT_ID1}
\end{equation}
\normalsize
where
\footnotesize
\begin{equation}
\mathcal{F}_k(x, y) = - \log \frac{e^{(S_y + \lambda_1 D_y)/\tau}}{\sum\limits_{i=1}^{|P|}e^{(S^p_i + \lambda_1 D^p_i)/\tau} + \sum\limits_{j=1}^{|N_k|} e^{(\lambda_2 (S^n_j + D^n_j))/\tau}}.\label{eq:PT_ID2}
\end{equation}
\normalsize
In Equation~\ref{eq:PT_ID2}, negative labels selected by Superclass-BG are partitioned into $K$ groups. Let $N_k$ denote the $k$th negative label group. Define the similarity between the image and the ground-truth positive label as $S_y = S(\mathcal{I}(x), \mathcal{T}(P_y))$ and dissimilarity between ground-truth positive label and ground-truth class label as $D_y = -D(\mathcal{T}(C_y), \mathcal{T}(P_y))$. Similarly, we also define the similarity between the image and $i$th positive label as $S^p_i = S(\mathcal{I}(x), \mathcal{T}(P_i))$ and the corresponding positive dissimilarity as $D^p_i = D(\mathcal{T}(P_y), \mathcal{T}(P_i)) - D(\mathcal{T}(C_y), \mathcal{T}(P_i))$. Additionally, we denote the similarity between the image and $j$-th negative label in group $k$ as $S^n_j = S(\mathcal{I}(x), \mathcal{T}(N_{k_j}))$ and the corresponding negative dissimilarity as $D^n_j = -D(\mathcal{T}(P_y), \mathcal{T}(N_{k_j}))$. Finally, $\lambda_1$ and $\lambda_2$ are weighting hyperparameters. The ID loss aims to enhance the alignment between positive label embeddings and image embeddings while preserving the robust generalization capabilities inherent in the pre-trained CLIP model. Additionally, it aims to maximize the distinction between the image embeddings and the most challenging group of negative labels, thereby improving the model's discriminative power in the worst-case scenario. Through the utilization of $S^p_i$ and $D^p_i$, we refine the semantic relationships between positive labels, encouraging intra-class cohesion while preserving some aspects of the original inter-class structure. Concurrently, $D^n_j$ further amplifies the distance between positive and negative labels. Conversely, the OOD loss, formalized in Equation~\ref{eq:PT_OOD1} and~\ref{eq:PT_OOD2}, is designed to minimize the overall similarity between OOD images $x$, sourced from $\mathcal{D}_{\mathrm{OOD}}$, and the positive labels, while explicitly exposing the model to background description generated through the LLMs.
\footnotesize
\begin{equation}
\mathcal{L}_{\mathrm{OOD}} = \mathbb{E}_{(x)\sim \mathcal{D}_{\mathrm{OOD}}}
\big[\mathcal{G}(x)\big],\label{eq:PT_OOD1}
\end{equation}
\normalsize
where
\footnotesize
\begin{equation}
\mathcal{G}(x) = \frac{e^{\sum\limits_{i=1}^{|P|} (S(\mathcal{I}(x), \mathcal{T}(P_i)) / \tau)}}{\sum\limits_{i=1}^{|P|} e^{(S(\mathcal{I}(x), \mathcal{T}(P_i)) / \tau)} + \sum\limits_{j=1}^{|SC|} e^{(S(\mathcal{I}(x), \mathcal{T}(BG(sc_j))) / \tau)}}.\label{eq:PT_OOD2}
\end{equation}
\normalsize
To construct our final training objective, we employ gradient-aware scaling before balancing the two loss components. Specifically, we scale the OOD loss to match the gradient magnitude of the ID loss, ensuring comparable influence on model updates.
\footnotesize
\begin{equation}
     \mathcal{L}_{\mathrm{total}} = \mathcal{L}_{\mathrm{scaled\_OOD}} + \lambda_3 \cdot \mathcal{L}_{\mathrm{ID}},
\end{equation}
\normalsize
where $\lambda_3$ is a balancing hyperparameter.
\subsubsection{Phase-2 Visual Prompt Tuning.}
In the first phase, we exclusively employ PT of the label space to enhance the ability of positive labels to distinguish ID and OOD samples while preserving the visual architecture of the model. This approach prompts an inquiry into potential OOD detection improvements achievable through visual space tuning. Post-acquisition of the refined positive labels via PT, we implement VPT to align the training images with these labels using cross entropy and OOD losses. The process initiates with a enhanced positive embedding derived from a weighted sum of the learned positive labels and class labels. This methodology leverages class labels to preserve the pretrained knowledge of CLIP, mitigating overfitting risks in few-shot training scenarios. Let \(x\) be the input image with a learnable visual prompt from a few-shot training dataset and \(y\) be the corresponding ground truth label, and $P'$ denote the enhanced positive embeddings. The cross entropy loss is formulated as
\footnotesize
\begin{equation}
    \mathcal{L} = \mathbb{E}_{(x,y) \sim \mathcal{D}_{\mathrm{train}}} 
    \left[-\log \frac{e^{S(\mathcal{I}(x), P_y')/\tau}}{\sum\limits_{i=1}^{|P'|} e^{S({\mathcal{I}(x), P_i'})/\tau}}\right].
\end{equation}
\normalsize
The design of the OOD loss for VPT aligns with the principles of PT, both aiming to minimize the similarity between OOD images and learned positive labels. The OOD loss for VPT is formulated as
\footnotesize
\begin{equation}
 \mathcal{L}_{\mathrm{VPT\_OOD}} = \mathbb{E}_{(x)\sim \mathcal{D}_{\mathrm{OOD}}}
\big[\mathcal{H}(x)\big],
\end{equation}
\normalsize
where
\footnotesize
\begin{equation} 
\mathcal{H}(x) = \frac{\max\limits_{i, 0 \leq i \leq |P'|} e^{(S(\mathcal{I}(x), P_i') / \tau)}}{\sum\limits_{i=1}^{|P'|} e^{(S(\mathcal{I}(x), P_i') / \tau)} + \sum\limits_{j=1}^{|SC|} e^{(S(\mathcal{I}(x), \mathcal{T}(BG(sc_j)) / \tau)}}.
\end{equation}
\normalsize
In this phase, we deal with a single learnable visual prompt while the positive labels are now fixed. With only one learnable prompt, optimizing against all positive labels simultaneously might be too constrained and potentially lead to conflicts. Given the current visual representation, the max operation allows the model to focus on the most critical case - the positive label most similar to the OOD sample. To formulate our final training objective for VPT, we employ a methodology analogous to that used in the prompt tuning phase.
\footnotesize
\begin{equation}
    \mathcal{L}_{\mathrm{VPT\_total}} = \mathcal{L} + \lambda_4 \cdot \mathcal{L}_{\mathrm{scaled\_VPT\_OOD}},
\end{equation}
\normalsize
where $\lambda_4$ is a balancing hyperparameter.
\begin{table*}[!t]
    \centering
    \setlength{\tabcolsep}{1.8pt}
    \small
    \begin{tabular}{@{}lccccccccccc@{}}
        \toprule
        \multirow{2}{*}{\textbf{Method}} & \multirow{2}{*}{\textbf{\#}} & \multicolumn{2}{c}{\textbf{iNaturalist}} & \multicolumn{2}{c}{\textbf{SUN}} & \multicolumn{2}{c}{\textbf{Places}} & \multicolumn{2}{c}{\textbf{Texture}} & \multicolumn{2}{c}{\textbf{Average}} \\ 
        \cmidrule(lr){3-4} \cmidrule(lr){5-6} \cmidrule(lr){7-8} \cmidrule(lr){9-10} \cmidrule(lr){11-12}
        & & FPR95 $\downarrow$ & AUROC$\uparrow$ & FPR95 $\downarrow$ & AUROC $\uparrow$ & FPR95 $\downarrow$ & AUROC $\uparrow$ & FPR95 $\downarrow$ & AUROC $\uparrow$ & FPR95 $\downarrow$ & AUROC $\uparrow$ \\
        \midrule
        \multicolumn{12}{c}{\textbf{zero-shot}} \\
        Mahalanobis \cite{lee2018simple} & 0 & 99.33 & 55.89 & 99.41 & 59.94 & 98.54 & 65.96 & 98.46 & 64.23 & 98.94 & 61.50 \\
        Energy \cite{liu2020energy} & 0 & 81.08 & 85.09 & 79.02 & 84.24 & 75.08 & 83.38 & 93.65 & 65.56 & 82.21 & 79.57 \\
        ZOC \cite{esmaeilpour2022zero} & * & 87.30 & 86.09 & 81.51 & 81.20 & 73.06 & 83.39 & 98.90 & 76.46 & 85.19 & 81.79 \\
        MCM \cite{ming2022delving} & 0 & 20.91 & 94.61 & 37.59 & 92.57 & 44.69 & 89.77 & 57.77 & 86.11 & 42.74 & 90.77 \\
        CLIPN \cite{wang2023clipn} & * & 23.94 & 95.27 & 26.17 & 93.93 & 33.45 & 92.28 & 40.83 & 90.93 & 31.10 & 93.10 \\
        NegLabel \cite{jiang2024neglabel} & 0 & 1.91 & 99.49 & 20.53 & 95.49 & 35.59 & 91.64 & 35.59 & 90.22 & 25.40 & 94.21 \\
        \rowcolor{gray!20}Ours (Superlcass) & 0 & 1.26 & 99.68 & 20.74 & 94.85 & 35.80 & 91.60 & 47.95 & 88.21 & 26.44 & 93.58 \\
        \rowcolor{gray!20}Ours (Superclass-BG) & 0 & 1.15 & 99.71 & 13.36 & 96.64 & 26.62 & 93.70 & 44.26 & 89.57 & 21.34 & 94.90 \\
        \midrule
        \multicolumn{12}{c}{\textbf{requires training}} \\
        NPOS \cite{tao2023nonparametric} & F & 16.58 & 96.19 & 43.77 & 90.44 & 45.27 & 89.44 & 46.12 & 88.80 & 37.93 & 91.22 \\
        LSN \cite{nie2024outofdistribution} & 64 & 21.56 & 95.83 & 23.62 & 94.35 & 34.48 & 91.25 & 38.54 & 90.42 & 30.22 & 92.96 \\
        ID-like \cite{bai2024idlike} & 4 & 8.98 & 98.19 & 42.03 & 91.64 & 44.00 & 90.57 & \underline{25.27} & \underline{94.32} & 30.07 & 93.68 \\
        LoCoOp \cite{miyai2023locoop} & 16 & 16.05 & 96.86 & 23.44 & 95.07 & 32.87 & 91.98 & 42.28 & 90.19 & 28.66 & 93.52 \\
        NegPrompt \cite{li2024learning} & 16 & 6.32 & 98.73 & 22.89 & 95.55 & 27.60 & 93.34 & 35.21 & 91.60 & 23.01 & 94.81 \\
        \rowcolor{gray!20} Ours (Train) & 4 & \textbf{0.65} & \textbf{99.83} & \textbf{12.07} & \underline{96.91} & \underline{23.21} & \underline{94.47} & 28.10 & 93.80 & \underline{16.01} & \underline{96.25}\\
        \rowcolor{gray!20} Ours (Train) & 16 & \underline{0.68} & \underline{99.81} & \underline{13.15} & \textbf{96.94} & \textbf{16.03} & \textbf{96.55} & \textbf{11.74} & \textbf{97.38} & \textbf{10.4} & \textbf{97.67}\\
        \bottomrule
    \end{tabular}
    \caption{Comparisons of the proposed method and competitive baselines on the ImageNet-1K dataset. The best result in each column is in bold, and the second-best is underlined. The values for Superclass and Superclass-BG are derived from the average of 3 independent generations using identical prompts. All methods are based on CLIP-B/16, which employs a ViT-B/16 as the image encoder and a masked self-attention Transformer as the text encoder. All values are percentages. Performance metrics for baseline methods are cited from NegLabel and their respective original publications. The \# column indicates the required number of samples per class for training, where F denotes full fine-tuning. The methods with $^*$ indicate the requirement of additional training. The shaded part represents our method.}
    \label{tab:Result}
\end{table*}
\subsubsection{Few-shot Inference}
Similar to zero-shot OOD setting, 
The negative labels, mined through the proposed Superclass-BG approach, are also partitioned into $K$ distinct groups during inference. We compute softmax probabilities for each group, incorporating both enhanced positive embeddings and the respective negative label group. The OOD detection score function aggregates probabilities assigned to enhanced positive embeddings across all groups. Formally, for input image $x$, enhanced positive label embeddings $P'$, negative label groups $N_k$, and total group count $K$, the score function $\alpha(x)$ is defined as:

\footnotesize
\begin{equation}
\alpha(x) = \sum_{i=1}^{K} \frac{\sum\limits_{i=1}^{|P'|} e^{(S(\mathcal{I}(x), P_i') / \tau)}}{\sum\limits_{i=1}^{|P'|} e^{(S(\mathcal{I}(x), P_j') / \tau)} + \sum\limits_{j=1}^{|N_k|} e^{(S(\mathcal{I}(x), \mathcal{T}(N_{k_j}) / \tau)}}.
\end{equation}
\normalsize
\section{Experiments}
\subsubsection{Datasets and Benchmarks.}
We utilize the ImageNet-1K OOD benchmark~\cite{huang2021importance} to compare our method with existing zero-shot and few-shot training-based OOD detection approaches. The ImageNet-1K OOD benchmark employs ImageNet-1K as the ID dataset and uses iNaturalist \cite{van2018inaturalist}, SUN \cite{xiao2010sun}, Places \cite{zhou2018places}, and Texture \cite{cimpoi2014describing} as OOD datasets, which have no class overlap with the ID dataset. For the more challenging OOD detection, we follow the settings of MCM \cite{ming2022delving}, which adopt ImageNet-10 and ImageNet-20 alternately as ID and OOD datasets. Beyond the aforementioned datasets, we evaluate our proposed method on OpenOOD V1.5 ImageNet-1K full-spectrum benchmark \cite{zhang2023openood, vaze2022openset}. These benchmarks introduce more challenging Near-OOD datasets, such as SSB-hard \cite{vaze2022openset} and NINCO \cite{bitterwolf2023ninco}, and also test robustness against covariate shifts.
\subsubsection{Implementation Details.}
We employ CLIP-B/16 as our backbone architecture, which uses a ViT-B/16 as the image encoder and a masked self-attention Transformer as the text encoder. For few-shot training, we randomly sample 16 images per class. We create 256 random crops per image to generate OOD datasets and select the bottom two crops based on label similarity. Claude 3.5 Sonnet \cite{Anthropic2024} generates 10 three-word background descriptions using a temperature of 0. Following NegLabel \cite{jiang2024neglabel}, we use the lowest $15\%$ similarity scores in nouns and adjectives as negative labels. Prompt tuning is performed for $200$ epochs, with batch size $256$ and learning rate $0.025$ (SGD). VPT is trained for 5 epochs, with batch size $32$ and learning rate $0.2$ (SGD). ID loss hyperparameters are $\lambda_1 = \lambda_2 = 0.25$. ID and OOD loss balancing weights are $\lambda_3 = \lambda_4 = 0.3$ %
Enhanced positive embeddings incorporate equal contributions $(0.5)$ for class labels and learned positive labels. All experiments are conducted on a single NVIDIA 4090 GPU. 
\subsubsection{Evaluation Metrics.}
We employ two primary metrics: (1) FPR95, which is the probability that an OOD example is misclassified as ID when the true positive rate is as high as 95\%; (2) Area Under the Receiver Operating Characteristic (AUROC).
\subsection{Results and Discussions}
\begin{table}[!t]
\setlength{\tabcolsep}{1.5pt}
\centering
\small
\begin{tabular}{lccccc}
\toprule
\multirow{2}{*}{\textbf{Method}} & \multirow{2}{*}{\textbf{\#}} & \multicolumn{2}{c}{\textbf{Near-OOD}} & \multicolumn{2}{c}{\textbf{Far-OOD}} \\
\cmidrule(lr){3-4} \cmidrule(lr){5-6} &
& FPR95$\downarrow$ & AUROC$\uparrow$ & FPR95$\downarrow$ & AUROC$\uparrow$ \\
\midrule
\multicolumn{5}{c}{\textbf{zero-shot}} \\
MCM  & 0 & 94.74 & 58.11 & 77.47 & 82.56 \\
NegLabel & 0 & 76.97 & 70.11 & \underline{28.80} & \underline{93.52} \\
\rowcolor{gray!20} Ours (Superclass-BG) & 0 &\underline{66.52} & 77.98 & 32.53 & 93.03 \\
\midrule
\multicolumn{5}{c}{\textbf{requires training}} \\
NNGuide & F & 74.28 & 71.85 & 34.28 & 92.24 \\
CoOp & 16 & 95.82 & 56.36 & 80.31 & 82.41 \\
CoCoOp & 16 & 96.23 & 53.11 & 79.59 & 74.48 \\
LoCoOp & 16 & 90.91 & 59.34 & 54.33 & 84.02 \\
LSA & 16 & 70.56 & \underline{78.22} & 48.06 & 86.85 \\
\rowcolor{gray!20} Ours (Train) &  16 & \textbf{51.35} & \textbf{83.04} & \textbf{17.36} & \textbf{95.73} \\
\bottomrule
\end{tabular}
\caption{Comparison of the OpenOOD V1.5 full-spectrum benchmark \cite{zhang2023openood, yang2022fsood} between competitive baseline, including and the current state-of-the-art, NNGuide \cite{park2023nearest}, LSA \cite{lu2023likelihoodaware} and CoCoOp \cite{zhou2022conditional}. Near-OOD is the average of SSB-hard and NINCO, while Far-OOD is the average of iNaturalist, Texture, and OpenImage-O \cite{wang2022vim}.}
\label{tab:OpenOOD}
\end{table}
\begin{table}[!t]
\centering
\small
\setlength{\tabcolsep}{1pt}
\begin{tabular}{ccccc}
\toprule
\textbf{ID Dataset} & \textbf{OOD Dataset} & \textbf{Method} & \textbf{FPR95}$\downarrow$ & \textbf{AUROC}$\uparrow$\\
\midrule
\multirow{3}{*}{\textbf{ImageNet-10}} & \multirow{3}{*}{\textbf{ImageNet-20}} & MCM & 
\underline{6.69} & \underline{98.45} \\
 & & NegLabel & 8.69 & 98.39 \\
 & & \cellcolor{gray!20} Superclass-BG & \cellcolor{gray!20} \textbf{6.26} & \cellcolor{gray!20} \textbf{98.70} \\
\midrule
\multirow{3}{*}{\textbf{ImageNet-20}} & \multirow{3}{*}{\textbf{ImageNet-10}} & MCM & 9.46 & 
\underline{98.33} \\
 & & NegLabel & \underline{8.51} & 98.27 \\
 & & \cellcolor{gray!20} Superclass-BG & \cellcolor{gray!20} \textbf{6.56} & \cellcolor{gray!20} \textbf{98.46} \\
\bottomrule
\end{tabular}
\caption{Zero-shot performance comparison of OOD detection methods on challenging OOD datasets ImageNet-10 and ImageNet-20.}
\label{tab:hardOOD}
\end{table}
\subsubsection{Main Results.}
Table \ref{tab:Result} presents a comprehensive comparison of our proposed method against existing OOD detection approaches. On the ImageNet OOD benchmark, our method achieves an average improvement of 2.8$\%$ in AUROC and a reduction of 12.6$\%$ in FPR95 compared to the current state-of-the-art. For the more challenging OpenOOD V1.5 full-spectrum benchmark, the results are shown in Table \ref{tab:OpenOOD}. Our method exhibits remarkable robustness to covariate shifts. In the Far-OOD detection scenarios, we observe a 2.2$\%$ increase in AUROC and a 9.4$\%$ decrease in FPR95. The improvement is even more pronounced in Near-OOD scenarios, with an 4.8$\%$ increase in AUROC and a 19.2$\%$ decrease in FPR95. Notably, our zero-shot Superclass-BG approach surpasses current training-based state-of-the-art methods on the ImageNet-1K OOD benchmark. It also performs superior on the challenging ImageNet-10 and ImageNet-20 benchmarks, as evidenced in Table \ref{tab:hardOOD}.
\subsubsection{Influence of the Quantity of Background Description on Zero-Shot Performance.}
\begin{figure}[!t]
    \centering
    \includegraphics[width=0.47\textwidth]{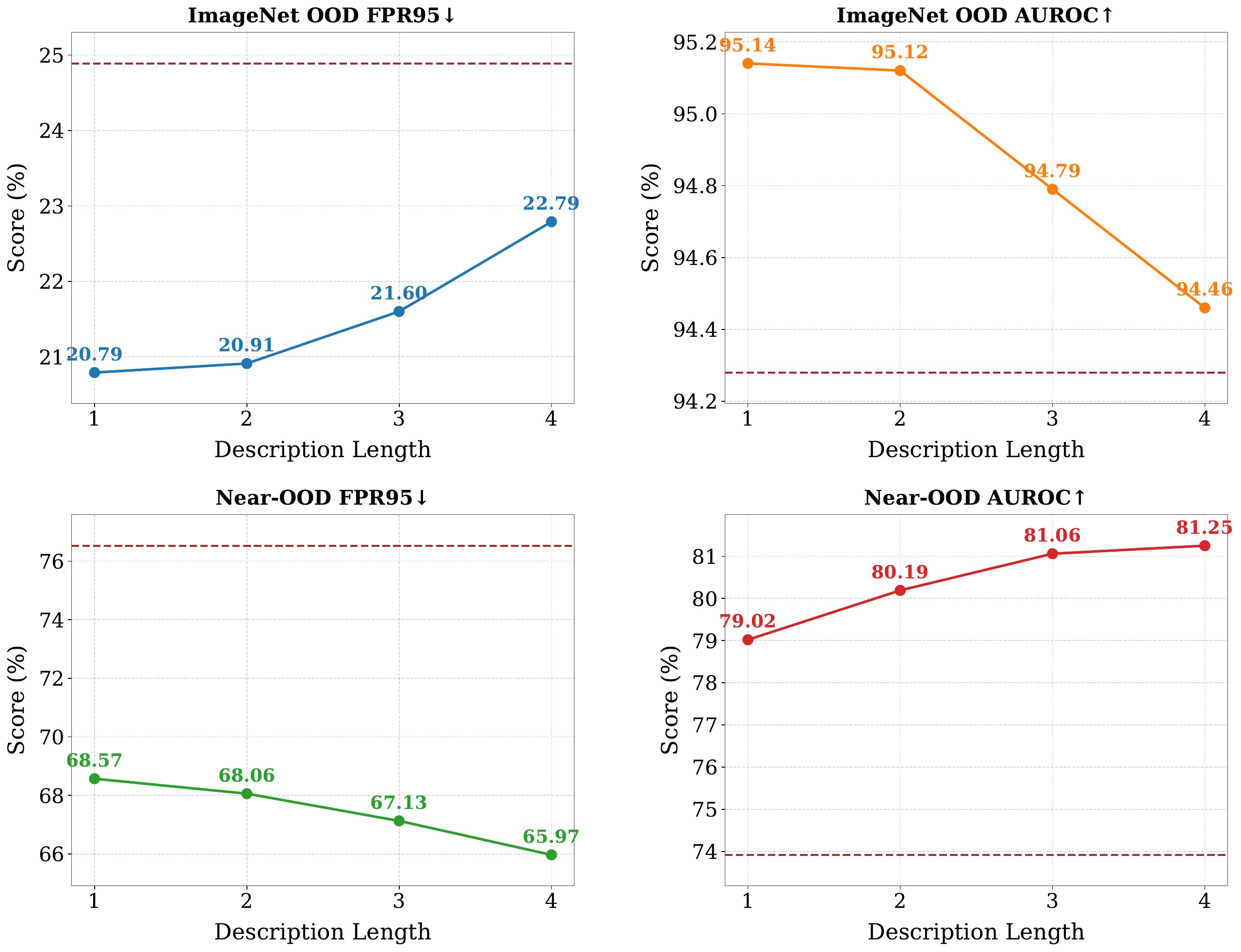}
    \caption{Superclass-BG zero-shot performance metrics as a function of description length. Each subplot shows the trend of a specific OOD detection metric for description lengths ranging from 1 to 4. Data points represent average scores computed from 5-15 descriptions. The brown dashed line indicates the performances of NegLabel \cite{jiang2024neglabel}.}
    \label{fig:description length}
\end{figure}
Figure \ref{fig:description length} reveals that shorter descriptions enhance performance on the ImageNet-1K OOD benchmark, while longer descriptions prove more effective for Near-OOD detection. We hypothesize that shorter descriptions, limited in capturing nuanced background features, may be oversimplified by omitting general characteristics. This approach benefits the simpler ImageNet-1K OOD benchmark but potentially removes possible indicators for distinguishing challenging Near-OOD samples. Conversely, removing broader background concepts might reduce overlap with straightforward OOD samples, contributing to improved performance on the ImageNet-1K OOD benchmark. On the other hand, longer descriptions provide a more comprehensive representation of both specific and general attributes, aiding in Near-OOD sample differentiation where similarities with ID data are shared. However, longer descriptions may retain general characteristics that overlap with straightforward OOD samples, presenting a trade-off in effectiveness across different OOD detection scenarios.
\subsubsection{Various Large Language Models.}
\begin{table}[!t]
\setlength{\tabcolsep}{2pt}
\centering
\small
\begin{tabular}{lcccc}
\toprule
\multirow{2}{*}{\textbf{LLMs}} & \multicolumn{2}{c}{\textbf{ImageNet OOD}} & \multicolumn{2}{c}{\textbf{Near-OOD}} \\
\cmidrule(lr){2-3} \cmidrule(lr){4-5}
& FPR95 $\downarrow$ & AUROC $\uparrow$ & FPR95 $\downarrow$ & AUROC $\uparrow$ \\
\midrule
Claude 3.5 Sonnet & \underline{21.34} & 94.90 & \underline{67.22} & 81.04 \\
GPT-4-Turbo & 21.39 & \underline{94.93} & \textbf{66.60} & \textbf{81.16} \\
GPT-4o & 22.07 & 94.62 & 67.44 & \underline{81.05} \\
GPT-4o-mini & \textbf{20.18} & \textbf{95.13} & 68.74 & 80.27 \\
\bottomrule
\end{tabular}
\caption{Superclass-BG zero-shot OOD detection performance comparison using various LLMs on ImageNet OOD and Near-OOD datasets. The results are the average of 3 independent generations using identical prompts.}
\label{tab:LLMs}
\end{table}
The results presented in Table \ref{tab:LLMs} show that utilizing smaller LLMs such as GPT-4o-mini \cite{OpenAI2024} yields performance similar to employing fewer words for background description generation. This phenomenon may be attributed to the limited capacity of these models to articulate nuanced features. Compared to GPT-4-Turbo, Claude 3.5 Sonnet demonstrates comparable performance while offering a more affordable solution with a faster generation speed. Consequently, we have opted to employ Claude 3.5 Sonnet for our experiments.
\subsubsection{Computational Complexity.} Training takes 194 minutes for phase 1 and 6 minutes for phase 2 on average, using a single NVIDIA 4090 GPU. Following the score function proposed by NegLabel, the computational complexity of our proposed method incurs about $\mathcal{O}(2Md)$ FLOPs extra computational burden per image during inference, where $M$ denotes the number of negative labels and $d$ represents the dimension of the embedding feature.
\begin{table}[!t]
\centering
\setlength{\tabcolsep}{5pt}
\small
\begin{tabular}{cccccc}
\toprule
\textbf{S-BG} & \textbf{Filter} & \textbf{PT} & \textbf{VPT} & \textbf{ImageNet OOD} & \textbf{Near-OOD} \\
\midrule
\multicolumn{6}{c}{\textbf{zero-shot}} \\
 $\times$& $\times$ & &  & 93.39 & 80.79 \\
\checkmark &  $\times$&  &  & 94.90 & 81.04 \\
 $\times$ & \checkmark &  &  & 94.28 & 73.92 \\
\checkmark & \checkmark &  &  & 95.51 & 72.52 \\
\midrule
\multicolumn{6}{c}{\textbf{few-shot}} \\
 $\times$& $\times$& \checkmark & $\times$ & 95.49 & 87.47 \\
 $\times$& $\times$& $\times$ & \checkmark & 94.60 & 76.06 \\
\checkmark &$\times$ & \checkmark & $\times$ & 96.63 & 87.96 \\
 $\times$& \checkmark & \checkmark & $\times$ & 96.39 & 83.24 \\
\checkmark & \checkmark & \checkmark & $\times$ & 97.33 & 80.34 \\
$\times$ & $\times$ & \checkmark &  \checkmark & 96.59 & \underline{91.25} \\
\checkmark & $\times$ & \checkmark & \checkmark & \underline{97.67} & \textbf{91.98} \\
 $\times$ & \checkmark & \checkmark & \checkmark & 97.25 & 87.43 \\
\checkmark & \checkmark & \checkmark & \checkmark & \textbf{97.77} & 82.49 \\
\bottomrule
\end{tabular}
\caption{Performance comparisons of various configurations in the proposed methods with ImageNet-1K as ID dataset. S-BG denotes Superclass-BG. Results are reported in terms of AUROC, with higher values indicating better performance.}
\label{tab:ablation}
\end{table}
\subsection{Ablation Study}
\subsubsection{The Effectiveness of Prompt Tuning and Visual Prompt Tuning.}
In Table \ref{tab:ablation}, we observe that few-shot prompt tuning yields performance improvements on both ImageNet OOD and Near-OOD datasets, regardless of whether Superclass-BG or filtering is employed. Notably, the enhancement is particularly pronounced for Near-OOD, with AUROC improvement of at least 6.6$\%$. Furthermore, the results demonstrate that the application of VPT in the second phase leads to additional performance gains.
\subsubsection{Pilot study of Candidate Label Filtering.}
When using ImageNet as the ID dataset, words from animal and food categories are excluded in NegLabel due to their prevalence classes; however, from the zero-shot and few-shot training results presented in Table \ref{tab:ablation}, we observe that filtered methods consistently outperform their unfiltered counterparts on the ImageNet OOD benchmark. Notably, the performance disparity is substantially more pronounced on Near-OOD datasets. We hypothesize that the slight decrease in performance on ImageNet OOD, observed when reincorporating candidate labels belonging to the major lexname category of the ID labels into the pool of selectable negative labels, may be attributed to the potential selection of negative labels semantically proximate to the major category. Conversely, the significant performance improvement on Near-OOD datasets can be explained by the inherent semantic proximity of Near-OOD samples to the ID semantic space. Consequently, these semantically related negative labels effectively serve as discriminative features for identifying Near-OOD samples. Additionally, in the absence of filtering, our proposed Superclass-BG method, which employs a more refined ID semantic space to select negative labels, further enhances performance on Near-OOD datasets.
\subsubsection{The Effectiveness of Enhancing Positive Labels with Class Labels.}
\begin{figure}[!t]
    \centering
    \includegraphics[width=0.47\textwidth]{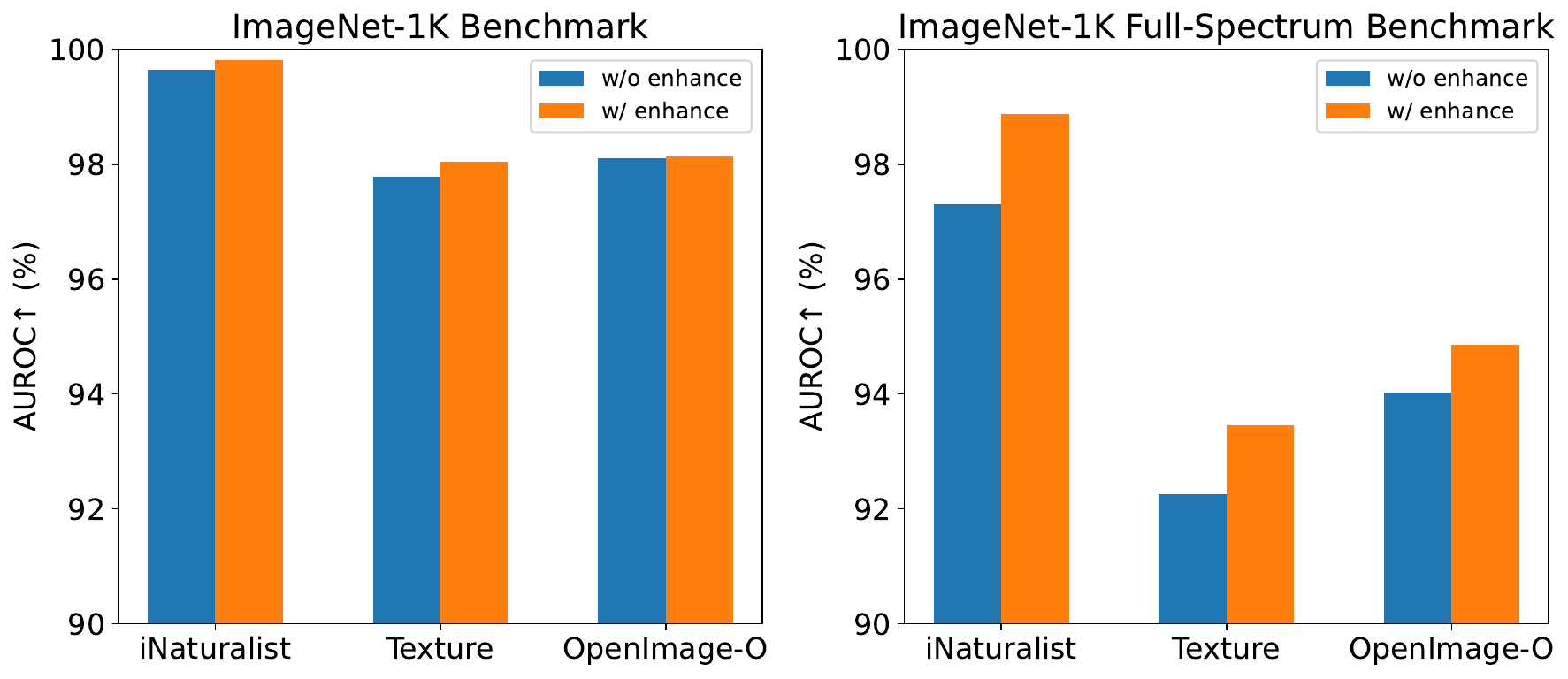}
    \caption{Performance comparison on the OpenOOD V1.5 benchmark \cite{zhang2023openood, yang2022fsood}. The evaluation contrasts the effectiveness of positive labels enhanced by original class labels against the baseline without enhancement.}
    \label{fig:enhance}
\end{figure}
As evidenced in Figure \ref{fig:enhance}, the OpenOOD V1.5 ImageNet-1K benchmark, which does not account for covariate shift, demonstrates that positive labels enhanced by class labels consistently outperform their non-enhanced counterparts across various datasets, albeit with marginal differences. However, we observe a significant amplification of this performance gap when examining the full-spectrum benchmark, which incorporates covariate shift robustness. This amplification underscores the efficacy of enhancing positive labels, adapted to the target distribution, with class labels that possess strong generalization capabilities.
\section{Conclusion}
In this work, we present a novel zero-shot OOD detection approach that harnesses the power of LLMs and VLMs. Our method introduces a hierarchical strategy using superclass labels and background descriptions generated by LLMs to select more representative negative labels from a comprehensive set of unfiltered candidate labels. Combining this approach with prompt tuning and visual prompt tuning, the proposed few-shot method further pushes the boundaries of OOD detection, consistently outperforming current state-of-the-art methods across multiple benchmarks, including challenging full-spectrum benchmark. Moreover, our method demonstrates remarkable robustness to covariate shifts, addressing a critical challenge in real-world scenarios.
\bibliography{aaai25}

\end{document}